\documentclass[letterpaper]{article}
\usepackage[preprint]{aaai2027}
\usepackage[hyphens]{url}
\usepackage{graphicx}
\usepackage{natbib}
\usepackage{caption}
\usepackage{booktabs}
\usepackage{amsmath}
\usepackage{amssymb}
\usepackage{xcolor}
\definecolor{codeblue}{HTML}{0B6FB0}

\newcommand{\SimNorm}{\operatorname{SimNorm}}
\newcommand{\zobs}{z^{\mathrm{obs}}}

\title{Fast Plasticity Recalibrates a Frozen World Model}
\author{ Sunwoo Hong\textsuperscript{\rm 1}, Chan Park\textsuperscript{\rm 2}, Ali Pesaranghader\textsuperscript{\rm 3} } 

\affiliations
{ \textsuperscript{\rm 1}KAIST AI, \textsuperscript{\rm 2}Sungkyunkwan University, \textsuperscript{\rm 3}LG Electronics, Toronto AI Lab\\[4pt] sunwoo@kaist.ac.kr
}

\begin{document}

\maketitle

\begin{abstract}
When a load is added to the arm or a muscle tires, the cerebellar adaptive-filter account explains motor recalibration through sensory prediction errors that adjust a readout over a stable representation. We bring this computation to a pretrained world-model controller whose body has changed. We call this \textbf{Fast Plasticity}: a small fast state recalibrates the frozen latent world model online. It is written once per real transition by a local prediction-error rule and read by every imagined transition in model predictive control. Nothing is backpropagated at test time and no pretrained weight changes. The central finding is how little state this requires. A latent bias of 512 coefficients recovers substantial control alongside a 64-fold larger state--action-keyed matrix. With one fixed update shared across tasks and evaluated on 20 held-out seeds, the bias raises Walker Walk return from 705 to 861 when actuation is 40\% weaker, Cheetah Run from 350 to 446 when it is 30\% weaker, and Hopper Hop from 243 to 296 after a 30\% increase in torso mass, where it improves on all 20 seeds. On identical transitions, earlier writes lower latent prediction error by 4--19\%, and fixed-candidate interventions show that the correction changes which plan the controller selects. Swimming marks the boundary of the method: there, large corrections improve prediction and lower return. Fast plasticity gives a pretrained internal model a direct path from recent sensory error to the next decision.
Code is available at \textcolor{codeblue}{\url{https://github.com/SunwooHong/fast-plasticity}}.
\end{abstract}

\begin{figure*}[t]
\centering
\includegraphics[width=\textwidth]{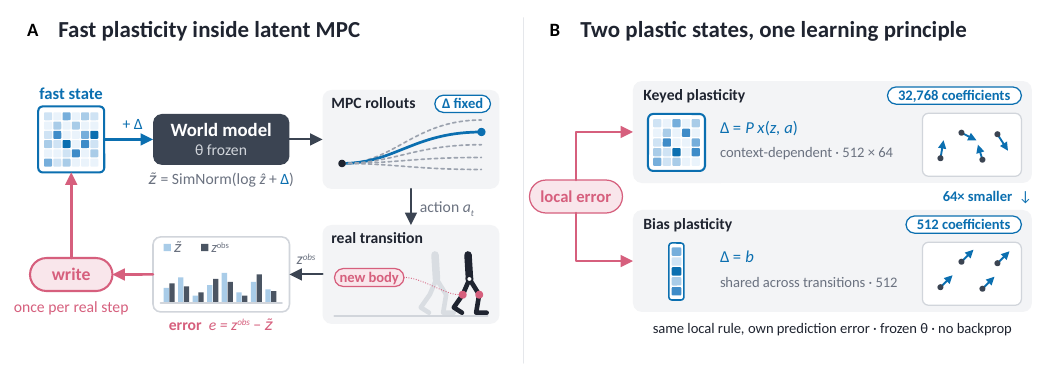}
\caption{Fast plasticity in a frozen world-model controller. \textbf{A:} A fast state adds a correction $\Delta$ to every imagined latent transition within an MPC planning call. After the real transition, the latent prediction error is written into the state once. Pretrained weights stay fixed. \textbf{B:} Two plastic states share one learning principle: each is written by the same local rule from its own prediction error. Keyed plasticity reads a state--action-dependent correction from a $512\times 64$ matrix. Bias plasticity reads one shared 512-dimensional correction, a 64-fold smaller state.}
\label{fig:mechanism}
\end{figure*}

\section{Introduction}

Lift a heavy bag or reach with a tired arm, and the first movements fall short. After a few repetitions they are accurate again. Internal-model theories explain this recovery. The brain predicts the sensory consequences of each motor command and uses the error between predicted and observed movement to recalibrate that prediction \citep{wolpert1998internal}. The cerebellar adaptive-filter account makes the computation concrete: a teaching error adjusts a small set of output weights while the input representation stays in place \citep{dean2010cerebellar}. State-space models of motor learning describe each such adaptive process with two quantities, an error sensitivity and a retention rate \citep{smith2006interacting}.

Pretrained world-model controllers have the stable representation and lack the fast readout. Their dynamics network encodes how actions change the world for the body it was trained on. When that body changes at deployment, through a weaker actuator or a heavier torso, the controller keeps planning with the old dynamics. Standard remedies retrain the network or prepare an adaptation module in advance. Yet a fresh prediction error is already available at every control step.

We give a frozen world model this missing pathway and ask how much control it restores. The pretrained networks play the role of the stable representation, and a small fast state plays the role of the adjustable readout. The fast state adds a correction to the latent transition of a frozen TD-MPC2 model \citep{hansen2024tdmpc2}. After each real transition, the latent prediction error is written into that state by a delta rule. During the next planning call, every imagined transition reads it (Figure~\ref{fig:mechanism}A). The encoder, dynamics, reward, value, and policy networks stay fixed, and no gradient passes through them.

The design question is where an error should be reused. \emph{Keyed plasticity} stores errors in a $512\times 64$ matrix addressed by state--action features, so a correction applies in the context where it was observed. \emph{Bias plasticity} stores them in one 512-dimensional vector, so a correction applies to every transition (Figure~\ref{fig:mechanism}B). Across three legged bodies and two kinds of physical change, both forms recover substantial control, and the 512-coefficient bias reaches a higher mean return than the 32,768-coefficient matrix in three of seven changed-dynamics conditions.

Our contributions are as follows.
\begin{itemize}
\item \textbf{A minimal interface for online recalibration.} One local write per real transition and one additive read per imagined transition, inserted into a frozen latent MPC controller with a single setting for all tasks.
\item \textbf{Control recovered under changed bodies, confirmed on held-out seeds.} Both forms improve return in Walker Walk, Hopper Hop, and Cheetah Run under weakened actuation, and in Hopper Hop under increased torso mass, on a 20-seed cohort run after the method was fixed.
\item \textbf{A 512-coefficient correction is enough.} The shared bias recovers substantial control with a 64-fold smaller state than the keyed matrix. Overlap among state--action keys and a constant-key reduction help explain why.
\pagebreak[4]
\item \textbf{An account of how the correction reaches behavior.} Planner interventions show that the fast state re-ranks \mbox{MPC} candidates, and a swimming task identifies the regime in which re-ranking lowers return.
\end{itemize}

\section{Related Work}

\paragraph{Internal models and local adaptation.}
Internal-model theories tie movement adaptation to sensory prediction error \citep{wolpert1998internal}. Cerebellar adaptive-filter models combine input features, adjustable coefficients, and a teaching error \citep{dean2010cerebellar}. State-space models of motor learning separate error sensitivity from retention \citep{smith2006interacting}. Our fast states implement these three roles inside a learned latent transition model: a read function, a signed latent error, and a leak.

\paragraph{Fast states in learning and control.}
Fast-weight models combine slow weights for long-term knowledge with rapidly changing state for recent experience \citep{ba2016fast}. Fast-weight programmers learn context-dependent writes \citep{schmidhuber1992fast,schlag2021linear}, and differentiable plasticity learns the coefficients governing plastic connections \citep{miconi2018differentiable}. In control, meta-learned adaptation \citep{nagabandi2019learning}, RMA \citep{kumar2021rma}, and Neural-Fly \citep{oconnell2022neuralfly} handle changing dynamics with representations trained for adaptation. Control-oriented meta-learning trains adaptive features for closed-loop tracking performance \citep{richards2023control}. Our update needs no such training. It attaches to an existing pretrained controller.

\paragraph{Model correction and latent planning.}
Offset-free MPC augments a nominal process model with integrating disturbance states to compensate for model mismatch \citep{pannocchia2003disturbance}. Gaussian-process MPC learns additive residual dynamics around a nominal model and accounts for their uncertainty during planning \citep{hewing2020cautious}. Learned latent dynamics and reward models support online planning in PlaNet and TD-MPC2 \citep{hafner2019planet,hansen2024tdmpc2}. DINO-WM learns action-conditioned dynamics over frozen pretrained visual features and plans toward image goals \citep{zhou2025dinowm}. Our fast state applies the correction directly in the frozen model's SimNorm coordinates, where it changes the latent trajectories scored by pretrained reward and value heads.

\paragraph{Adapting world models at deployment.}
Several recent methods adapt a pretrained latent world model to changed dynamics. ReDRAW trains residual corrections of latent-state dynamics from target-environment data \citep{lanier2026redraw}. AdaJEPA fine-tunes parts of a JEPA world model with gradient steps inside the MPC loop \citep{wang2026adajepa}, and JEPA-TTT trains the latent predictor persistently across episodes from a replay buffer \citep{zhang2026jepattt}. Sandwich-Residuals keeps the pretrained weights fixed and learns small corrections around the predictor \citep{soni2026sandwich}, and ALRA fits a 210-parameter affine residual on the action embedding \citep{soni2026alra}. These methods use gradient-based optimization for adaptation. Our write is a closed-form local update with no backpropagation through the network and no optimizer state. It runs once per real transition and corrects the SimNorm latent that MPC reads.

\paragraph{Representation-level control.}
In vision-language-action policies, learned linear observers and minimum-norm activation interventions steer behavior while the pretrained model remains frozen \citep{buurmeijer2026observing}. Our fast state applies transition-error feedback to the latent dynamics scored by MPC.

\section{Fast Plasticity in Latent MPC}

\subsection{The Correction Interface}

Let $z_t=h_\theta(o_t)\in\mathbb{R}^{512}$ be the latent encoding of observation $o_t$ and $a_t\in[-1,1]^{d_a}$ the command issued by the controller. The frozen dynamics predict $\hat z_{t+1}=f_\theta(z_t,a_t)$. TD-MPC2 groups latent coordinates in blocks of eight and applies SimNorm, a softmax within each group. We add a correction $\Delta_t$ in log coordinates:
\begin{equation}
\tilde z_{t+1}=\SimNorm\!\left(\log \hat z_{t+1}+\Delta_t(z_t,a_t)\right).
\label{eq:read}
\end{equation}
The corrected latent stays group-normalized, and $\Delta_t=0$ recovers the pretrained transition. After the real transition we encode the observation, $\zobs_{t+1}=h_\theta(o_{t+1})$, and compute the signed error
\begin{equation}
e_t=\zobs_{t+1}-\tilde z_{t+1}.
\label{eq:error}
\end{equation}
All parameters $\theta$ stay fixed during deployment. The fast state starts at zero in each episode.

\subsection{Bias Plasticity}

The bias state $b_t\in\mathbb{R}^{512}$ is the correction itself, $\Delta_t=b_t$. Its write integrates the error with a leak:
\begin{equation}
b_{t+1}=\Pi_c\!\left[(1-\lambda)\,b_t+\eta\,e_t\right],
\label{eq:bias}
\end{equation}
where $\Pi_c$ projects onto the norm ball of radius $c$. When the norm constraint is inactive, the update unrolls as
\begin{equation}
b_t=\eta\sum_{s<t}(1-\lambda)^{t-1-s}e_s .
\label{eq:bias-unrolled}
\end{equation}
The state is an exponentially weighted sum of recent errors, applied to every transition. The vector is the same for every transition, but its effect on the prediction is not. Within each SimNorm group, Eq.~\eqref{eq:read} gives $\tilde z_i\propto \hat z_i\exp(b_i)$, a multiplicative reweighting whose effect depends on the current prediction $\hat z$ and therefore on state and action. The planner composes this reweighting over the rollout horizon, and nonlinear reward and value heads score the result, so one vector shifts different candidate sequences by different amounts. Its 512 FP32 coefficients occupy 2~KiB.

\subsection{Keyed Plasticity}

The keyed state $P_t\in\mathbb{R}^{512\times 64}$ makes the correction depend on context. We form $u_t=[z_t^\top,a_t^\top]^\top$ and apply a fixed random projection with normalized Gaussian columns:
\begin{equation}
x_t=\frac{R^\top u_t}{\max(\lVert u_t\rVert_2,1)},\qquad R\in\mathbb{R}^{(512+d_a)\times 64}.
\label{eq:key}
\end{equation}
The read is $\Delta_t=P_t x_t$ and the write is a normalized delta rule:
\begin{equation}
P_{t+1}=\Pi_c\!\left[(1-\lambda)P_t+\eta\,\frac{e_t x_t^\top}{\lVert x_t\rVert_2^2+\epsilon}\right].
\label{eq:keyed}
\end{equation}
When the norm constraint is inactive, the update unrolls as
\begin{equation}
P_t x=\eta\sum_{s<t}(1-\lambda)^{t-1-s}e_s\,\frac{x_s^\top x}{\lVert x_s\rVert_2^2+\epsilon}.
\label{eq:keyed-unrolled}
\end{equation}
Comparing Eq.~\eqref{eq:keyed-unrolled} with Eq.~\eqref{eq:bias-unrolled} isolates the single difference between the two forms. The keyed read weights each past error by the similarity of its key to the current one. The bias read weights all past errors equally. With a constant key the keyed rule reduces to the bias rule (Appendix~\ref{app:constkey}). The matrix holds 32{,}768 coefficients, or 128~KiB.

\subsection{Local Learning and the Planning Read}

Both writes are gradient steps on a local objective. With the encoded target and key held fixed, soft-target cross-entropy through SimNorm satisfies
\begin{equation}
\frac{\partial L_{\mathrm{CE}}}{\partial \Delta_t}=-e_t,\quad
\nabla_{P_t}L_{\mathrm{CE}}=-e_t x_t^\top,\quad
\nabla_{b_t}L_{\mathrm{CE}}=-e_t .
\label{eq:local}
\end{equation}
Here $L_{\mathrm{CE}}=-\sum_{g=1}^{64}\sum_{i\in g}\zobs_{t+1,i}\log\tilde z_{t+1,i}$ sums over the 64 SimNorm groups of eight. The keyed update shares the error--feature outer-product structure of delta-rule fast-weight memories \citep{schlag2021linear}. The update therefore needs one vector addition or one outer product per real transition. It keeps no optimizer state and no transition buffer. The leak $\lambda=0.001$ gives a contribution a half-life of about 693 transitions, and $\eta$ sets the error sensitivity.

MPC reads the current correction in every imagined transition and holds the fast state fixed within a planning call. It then executes one action, scores the resulting prediction, and performs one write. The correction acts on behavior through the frozen reward and value heads, which score the corrected latents. For candidate action sequences $A$ and $B$ with scores $J_\Delta$ under correction and $J_0$ without,
\begin{equation}
J_\Delta(A)-J_\Delta(B)=J_0(A)-J_0(B)+\delta_\Delta(A)-\delta_\Delta(B),
\label{eq:rank}
\end{equation}
where $\delta_\Delta(\cdot)=J_\Delta(\cdot)-J_0(\cdot)$. A correction influences planning through its effect on relative candidate scores, which can change candidate rankings and the resulting MPPI-based action distribution \citep{williams2017mppi,hansen2024tdmpc2}. We measure this effect directly in Section~\ref{sec:planner}.

\section{Experimental Setup}

\paragraph{Controllers.}
We use the 5M-step, state-input TD-MPC2 checkpoints for the DeepMind Control Suite \citep{hansen2024tdmpc2,tassa2018dmcontrol}. The main tasks are Walker Walk, Hopper Hop, and Cheetah Run, three legged bodies with separate pretrained controllers. Fish Swim and Reacher Hard probe the method outside legged locomotion. Primary experiments use official checkpoint~1 for each task. Each episode has 500 controller decisions with action repeat two. The native planner has a three-decision horizon.

\paragraph{Body changes.}
The main change scales all actuators by a gain $g\in\{1.0,0.7,0.6\}$. The simulator receives $\mathrm{clip}(g\,a_t,-1,1)$ while the world model receives the original command $a_t$. A second change multiplies torso mass by 1.3. Additional tests assign different gains to individual joints. Every change alters the physical dynamics and leaves the task reward intact.

\paragraph{One fixed update.}
Both forms use $\eta=0.05$, $\lambda=0.001$, $\epsilon=10^{-6}$, and $c=10$ on every task, with one write per observed transition. These values were set during development on Walker and never changed.

\begin{table*}[t]
\centering
\small
\begin{tabular}{llrrrrr}
\toprule
Task & Gain & Frozen & Keyed & Bias & Keyed $-$ frozen [95\% CI] & Bias $-$ frozen [95\% CI] \\
\midrule
Walker Walk & 1.0 & 985.31 & 985.02 & 984.90 & $-0.28$ [$-0.84$, $0.28$] & $-0.41$ [$-1.09$, $0.28$] \\
 & 0.7 & 957.64 & 973.45 & 972.14 & $\mathbf{15.81}$ [$9.24$, $22.39$] & $\mathbf{14.50}$ [$8.22$, $20.79$] \\
 & 0.6 & 704.50 & 839.85 & 861.29 & $\mathbf{135.35}$ [$59.86$, $210.84$] & $\mathbf{156.79}$ [$88.46$, $225.13$] \\
\midrule
Hopper Hop & 1.0 & 567.02 & 569.44 & 569.80 & $2.41$ [$-0.07$, $4.89$] & $2.77$ [$-0.18$, $5.72$] \\
 & 0.7 & 6.72 & 15.62 & 14.17 & $\mathbf{8.90}$ [$5.07$, $12.72$] & $\mathbf{7.45}$ [$2.40$, $12.50$] \\
 & 0.6 & 4.58 & 5.10 & 4.73 & $0.52$ [$-0.05$, $1.10$] & $0.15$ [$-1.07$, $1.37$] \\
\midrule
Cheetah Run & 1.0 & 855.55 & 864.95 & 847.39 & $9.40$ [$-0.92$, $19.73$] & $-8.16$ [$-24.81$, $8.49$] \\
 & 0.7 & 350.11 & 403.99 & 445.88 & $\mathbf{53.88}$ [$11.71$, $96.04$] & $\mathbf{95.76}$ [$64.57$, $126.96$] \\
 & 0.6 & 236.37 & 266.60 & 257.13 & $\mathbf{30.23}$ [$11.61$, $48.84$] & $20.75$ [$-2.37$, $43.88$] \\
\midrule
Hopper Hop, mass $\times 1.3$ & 1.0 & 242.75 & 287.73 & 295.54 & $\mathbf{44.98}$ [$34.49$, $55.48$] & $\mathbf{52.80}$ [$43.00$, $62.59$] \\
\bottomrule
\end{tabular}
\caption{Return on the held-out confirmation cohort (seeds 20--39, checkpoint~1, 20 paired episodes per row). The same update setting is used in every row. Bold differences have 95\% intervals above zero.}
\label{tab:main}
\end{table*}

\begin{figure*}[t]
\centering
\includegraphics[width=\textwidth]{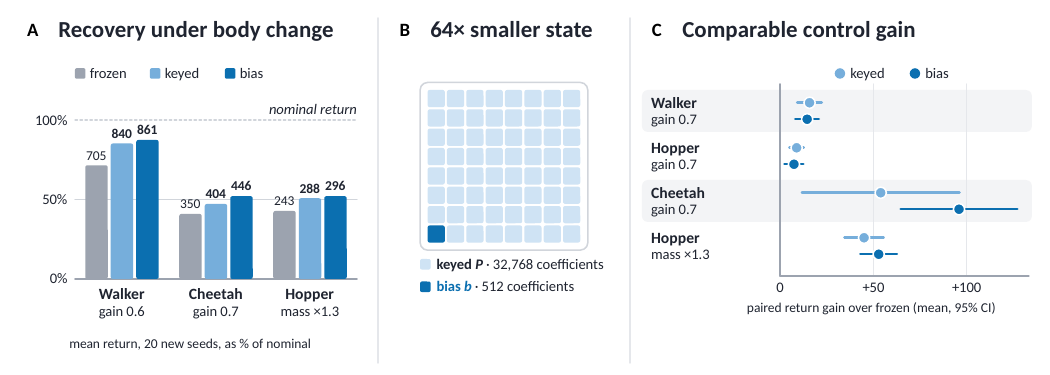}
\caption{A small shared correction restores control. \textbf{A:} Mean return on 20 held-out seeds as a percentage of nominal return for the three largest recoveries. Numbers above bars are raw returns. \textbf{B:} Fast-state size drawn to scale. Each cell stands for 512 coefficients. The keyed matrix fills all 64 cells and the bias fills one. \textbf{C:} Paired return gain over the frozen controller (mean and 95\% interval, 20 seeds) for every gain-0.7 condition and the Hopper mass change.}
\label{fig:results}
\end{figure*}

\paragraph{Held-out confirmation cohort.}
Development used evaluation seeds 0--19. After fixing the method we ran seeds 20--39 with identical settings. All headline results come from this confirmation cohort. Each comparison pairs a frozen and an adapted episode that share the environment and planner seed, and we report the mean paired return difference with a 95\% Student-$t$ interval over the 20 seeds. Every evaluated seed is retained.

\paragraph{Prediction on identical transitions.}
Along the adapted controller's trajectory we evaluate Eq.~\eqref{eq:read} twice before each write, once with the current fast state and once with zero correction. Target, state, and action are identical in the two reads, so the difference in squared error measures what earlier writes contributed to this prediction.

\section{Fast Plasticity Restores Control}

\subsection{Weakened Actuation}

Table~\ref{tab:main} reports every global-gain condition on the confirmation cohort. With 30\% weaker actuation, both forms improve return on all three bodies, and all six intervals lie above zero. On Walker the keyed and bias gains are 15.81 and 14.50 points, positive on 19 and 18 of 20 seeds. On Cheetah, frozen return drops from 856 to 350, and plasticity restores it to 404 (keyed) and 446 (bias).

The benefit grows with the disruption. At gain 0.6 the frozen Walker falls from 985 to 705. Bias plasticity brings it back to 861 and keyed plasticity to 840, recovering 56\% and 48\% of the lost return (Figure~\ref{fig:results}A). The fast state itself tracks the size of the mismatch: the mean keyed norm on Walker is 0.48 at nominal gain, 2.27 at gain 0.7, and 3.91 at gain 0.6, with the same ordering on Hopper and Cheetah. The norm cap is inactive in every locomotion run.

A uniform gain of 0.7 collapses Hopper to its reward floor, from 567 to 6.7. Plasticity doubles what remains (15.6 and 14.2), and at gain 0.6 hopping is lost in all conditions. The mass change below gives Hopper a regime with room to recover.

\subsection{Increased Body Mass}

Raising Hopper's torso mass by 30\% lowers frozen return to 243. Keyed and bias plasticity reach 288 and 296, paired gains of 45.0 and 52.8 points with intervals of $[34.5, 55.5]$ and $[43.0, 62.6]$. Both forms improve on all 20 seeds. The development cohort gave the same result, with gains of 49.0 and 42.8. Fast plasticity therefore corrects a change in body inertia as well as a change in actuation, with no change to the update.

The same mass change barely affects Walker, whose frozen return stays at 985, so there is nothing to recover. Cheetah's response to added mass varies widely across seeds, with intervals that span zero for both forms (Appendix~\ref{app:mass}).

\subsection{Behavior Under Nominal Dynamics}

With an unchanged body the fast states have little to correct. At nominal gain, keyed return differences are $-0.28$, $2.41$, and $9.40$ points on Walker, Hopper, and Cheetah, and bias differences are $-0.41$, $2.77$, and $-8.16$. All six intervals include zero. The bias ends each episode with a norm of 0.13 to 0.28, no episode reaches the norm cap, and same-transition prediction changes by $10^{-7}$ to $10^{-6}$ in MSE. Reacher Hard stays near its ceiling under every change we applied, and both forms leave its return within 0.2 points of frozen (Appendix~\ref{app:reacher}).

\section{A 512-Coefficient Correction Is Enough}
\label{sec:small}

Both forms recover substantial control performance, despite a 64-fold difference in adaptive-state size (Figure~\ref{fig:results}B--C). At gain 0.7 the bias tracks the keyed matrix on Walker and Hopper and exceeds it on Cheetah by 42 points. Under the Hopper mass change it is ahead by 8 points. At gain 0.6 the bias leads on Walker and the keyed state leads on Cheetah. Across the seven changed-dynamics rows of Table~\ref{tab:main}, the bias has the higher mean in three and the keyed state in four, and their intervals overlap in every row.

Three observations help explain why a shared correction goes this far.

\paragraph{Keys overlap.}
Over 200 normalized state--action keys collected along controller trajectories, mean pairwise cosine similarity is 0.69 on Walker and 0.60 on Cheetah. By Eq.~\eqref{eq:keyed-unrolled}, most of each stored error is then reused at every other key, which is what the bias does by construction. Part of this overlap comes from the key itself. SimNorm latents are non-negative, and an uncentered random projection of non-negative vectors produces keys with a large common component. Analyses of sensory representations likewise connect expansion and sparsity to pattern overlap \citep{babadi2014sparseness}. We therefore also tested a sparse high-dimensional key: a 512-dimensional ReLU projection that keeps its 51 largest coordinates. On Walker at gain 0.7 it gains 8.5 points, against 9.8 for the 64-dimensional key and 11.9 for the bias (development seeds 0--19). Sparse expansion alone leaves the comparison unchanged in this setting.

\paragraph{A constant key reproduces the bias.}
Fixing the key turns the keyed rule into the bias rule analytically (Appendix~\ref{app:constkey}). Empirically the three controllers line up: on Walker at gain 0.7 in the development cohort, keyed, constant-key, and bias plasticity return 969.5, 970.9, and 971.6 against 959.7 for frozen.

\paragraph{Integration is the essential ingredient.}
Carrying forward only the most recent residual, with no leaky accumulation, is harmful: it lowers return by 209 points on Walker at gain 0.7 and by 135 on Cheetah at gain 0.6. A diagonal Kalman bias, another small integrating state, gains 14.9 and 26.9 points on the same two settings (Appendix~\ref{app:controls}). What matters is an integrated estimate of the recent error. Its dimension can be as small as the latent itself.

A shared correction is sufficient for several of the tested changes. Context-dependent capacity becomes visible in prediction under joint-specific gains, where the keyed state yields larger same-transition improvements than the bias on Walker. Return is the same within one point there, with keyed-minus-bias differences of $-0.73$ and $+0.57$ points for the two gain patterns.

\begin{table}[t]
\centering
\small
\begin{tabular}{llrr}
\toprule
Task & Gain & Keyed & Bias \\
\midrule
Walker Walk & 0.7 & 6.7\% & 5.5\% \\
 & 0.6 & 7.9\% & 6.9\% \\
Hopper Hop & 0.7 & 6.1\% & 4.3\% \\
 & 0.6 & 8.8\% & 7.1\% \\
Cheetah Run & 0.7 & 15.9\% & 15.6\% \\
 & 0.6 & 19.1\% & 17.1\% \\
\bottomrule
\end{tabular}
\caption{Relative reduction in pre-write latent MSE when the current fast state replaces zero correction on identical transitions (confirmation cohort, mean over seeds). Each form is scored on its own executed trajectories.}
\label{tab:pred}
\end{table}

\section{How the Correction Reaches the Planner}
\label{sec:planner}

\subsection{Earlier Writes Improve the Current Prediction}

Table~\ref{tab:pred} reports identical-transition scoring. In every task and at both shifted gains, the fast state lowers the error of the prediction made before the next write. Reductions range from 4.3\% to 19.1\%, are largest on Cheetah, and grow as the gain decreases. Since target, state, and action are held fixed, the improvement is attributable to the accumulated writes alone.

\subsection{The Correction Changes the Selected Plan}

A better transition model matters for control only through the planner's choice. To measure that choice we saved 100 decision states from five Walker runs at gain 0.7 and scored the same MPPI candidate sets twice, with the current keyed state and with zero correction. The top-ranked candidate changes at 89 of the 100 states. The elite-weighted first action changes at all 100, by a mean $\ell_2$ distance of 0.119. The fast state is an active input to action selection at nearly every step.

\subsection{Where Re-Ranking Hurts: Swimming}

Fish Swim shows the limit of the method. Under a uniform gain of 0.6, both forms improve same-transition prediction, and paired return changes are $-67.6$ $[-129.5, -5.8]$ for the keyed state and $-93.1$ $[-157.6, -28.6]$ for the bias. The keyed norm cap is active in all 20 episodes. Fish is the only task in which the correction ever reaches the cap. Large latent corrections change the trajectories evaluated by the frozen reward and value heads. The Fish results show that improved local prediction can alter these planning computations without improving closed-loop control.

Correction strength shapes the outcome. Under alternating joint gains the correction stays small, the cap is never reached, and keyed plasticity improves Fish return by 16 points ($[6,26]$). In a development sweep at gain 0.7, reducing the read strength to one quarter raises mean return from 633 to 655 while cutting the prediction improvement by more than half (Appendix~\ref{app:fish}). The legged gain experiments stay in the small-correction regime, with the cap never reached, and there prediction and control improve together.

\section{Relation to Explicit Gain Models}

An estimator that searches over scalar actuator gains has the true perturbation inside its hypothesis class, which makes it the natural reference on uniform-gain tests. Table~\ref{tab:gain} compares it with fast plasticity on the development cohort. The estimator is ahead on Walker at gain 0.7 (977 against 972 for the bias) and on Cheetah at gain 0.6 (285 against 252 for the keyed state). Fast plasticity is ahead on Hopper at gain 0.7 (23.5 against 16.4) and level on Cheetah at gain 0.7 (430 against 426).

The two approaches buy their performance differently. The estimator evaluates seven gain hypotheses over a 25-transition window and adds about 50~ms to each decision, where planning itself takes about 30~ms. It has no hypothesis for a change in mass. Fast plasticity assumes no perturbation family, stores no window, adds one vector to each imagined transition, and handles both the gain and the mass change with the same rule.

\begin{table}[t]
\centering
\small
\setlength{\tabcolsep}{4pt}
\begin{tabular}{llrrrrr}
\toprule
Task & Gain & Frozen & Keyed & Bias & Est. & Known \\
\midrule
Walker & 0.7 & 959.7 & 969.5 & 971.6 & 977.5 & 978.7 \\
Hopper & 0.7 & 9.7 & 23.5 & 17.4 & 16.4 & 19.9 \\
 & 0.6 & 6.0 & 5.9 & 5.5 & 8.6 & 9.3 \\
Cheetah & 0.7 & 381.9 & 419.9 & 429.5 & 426.2 & 424.5 \\
 & 0.6 & 232.8 & 252.5 & 249.0 & 285.2 & 283.9 \\
\bottomrule
\end{tabular}
\caption{Mean return against explicit gain models on development seeds 0--19. \emph{Est.} selects a scalar gain online from the last 25 transitions. \emph{Known} gives the dynamics model the true gain.}
\label{tab:gain}
\end{table}

\section{Discussion}

Fast plasticity turns the adaptive-filter account of sensorimotor learning into a working component of a modern latent planner. The pretrained networks hold the long-term knowledge of the task. A teaching error sets the direction of each update, a leak sets its persistence, and the read function sets where it applies \citep{dean2010cerebellar,smith2006interacting}. With these three pieces and no training, a frozen controller regains a large share of the return it loses when its body changes.

The main empirical lesson concerns the size of the adaptive state. A single shared correction of 512 coefficients restores a large share of the lost control. This has a practical consequence and a scientific one. Practically, online adaptation of a world-model controller fits in 2~KiB and one vector addition per imagined step, and the state stays near zero when the body is unchanged. Scientifically, it shows that for several body changes, a single integrating readout over a pretrained latent space is sufficient for effective recalibration.

The planner determines the value of a correction. Across the legged bodies, corrections stay small, prediction improves, candidate rankings change at most decisions, and return rises. Section~\ref{sec:planner} shows that larger corrections reshape planning in other ways, which makes the read strength a natural control variable. Selective supervision offers a related principle: Pivot-SD uses information gain over the remaining masked tokens to select denoising commitments for offline self-distillation \citep{kim2026pivotsd}. A corresponding extension here is to gate each read by its effect on candidate-plan rankings.

The read function is the second lever. The keyed state here addresses its matrix with a fixed random projection of the latent. The design of a sparse expanded key \citep{babadi2014sparseness} offers further control over which transitions share a correction, particularly under joint-specific changes.

\paragraph{Scope.}
Experiments use state-input TD-MPC2 in simulation. Additional pretrained checkpoints for Hopper and Cheetah give positive mean gains at gain 0.7 in 23 of 24 ten-seed blocks (Appendix~\ref{app:ckpt}).

\section{Conclusion}

A frozen world model adapts to a changed body through one local write per real step and one additive read per imagined step. On held-out seeds this restores control in three legged bodies under weakened actuation and in Hopper under added mass, with a single setting and no test-time backpropagation. A 512-coefficient bias recovers substantial control with a state 64 times smaller than the keyed matrix. Recent sensory error, integrated in a state the size of the latent, is enough to bring a pretrained internal model back in line with the body it controls.

\clearpage
\bibliography{fastplasticity}

\begin{thebibliography}{26}
\providecommand{\natexlab}[1]{#1}

\bibitem[{Ba et~al.(2016)Ba, Hinton, Mnih, Leibo, and Ionescu}]{ba2016fast}
Ba, J.; Hinton, G.~E.; Mnih, V.; Leibo, J.~Z.; and Ionescu, C. 2016.
\newblock Using Fast Weights to Attend to the Recent Past.
\newblock In \emph{Advances in Neural Information Processing Systems},
  volume~29, 4331--4339.

\bibitem[{Babadi and Sompolinsky(2014)}]{babadi2014sparseness}
Babadi, B.; and Sompolinsky, H. 2014.
\newblock Sparseness and Expansion in Sensory Representations.
\newblock \emph{Neuron}, 83(5): 1213--1226.

\bibitem[{Buurmeijer et~al.(2026)Buurmeijer, {Amo Alonso}, Swann, and
  Pavone}]{buurmeijer2026observing}
Buurmeijer, H.; {Amo Alonso}, C.; Swann, A.; and Pavone, M. 2026.
\newblock Observing and Controlling Features in Vision-Language-Action Models.
\newblock \emph{arXiv preprint arXiv:2603.05487}.

\bibitem[{Dean et~al.(2010)Dean, Porrill, Ekerot, and
  J{\"o}rntell}]{dean2010cerebellar}
Dean, P.; Porrill, J.; Ekerot, C.-F.; and J{\"o}rntell, H. 2010.
\newblock The Cerebellar Microcircuit as an Adaptive Filter: Experimental and
  Computational Evidence.
\newblock \emph{Nature Reviews Neuroscience}, 11(1): 30--43.

\bibitem[{Hafner et~al.(2019)Hafner, Lillicrap, Fischer, Villegas, Ha, Lee, and
  Davidson}]{hafner2019planet}
Hafner, D.; Lillicrap, T.; Fischer, I.; Villegas, R.; Ha, D.; Lee, H.; and
  Davidson, J. 2019.
\newblock Learning Latent Dynamics for Planning from Pixels.
\newblock In \emph{Proceedings of the 36th International Conference on Machine
  Learning}, volume~97 of \emph{Proceedings of Machine Learning Research},
  2555--2565. PMLR.

\bibitem[{Hansen, Su, and Wang(2024)}]{hansen2024tdmpc2}
Hansen, N.; Su, H.; and Wang, X. 2024.
\newblock {TD-MPC2}: Scalable, Robust World Models for Continuous Control.
\newblock In \emph{International Conference on Learning Representations}.

\bibitem[{Hewing, Kabzan, and Zeilinger(2020)}]{hewing2020cautious}
Hewing, L.; Kabzan, J.; and Zeilinger, M.~N. 2020.
\newblock Cautious Model Predictive Control Using Gaussian Process Regression.
\newblock \emph{IEEE Transactions on Control Systems Technology}, 28(6):
  2736--2743.

\bibitem[{Kim et~al.(2026)Kim, Hong, Choi, Chao, Yun, and
  Krishnan}]{kim2026pivotsd}
Kim, S.~H.; Hong, S.; Choi, Y.; Chao, C.-H.; Yun, S.-Y.; and Krishnan, R.~G.
  2026.
\newblock {Pivot-SD}: Efficient Self-Distillation for Masked Diffusion Language
  Models.
\newblock In \emph{Proceedings of the 2026 Conference on Empirical Methods in
  Natural Language Processing}, 48790--48806. Budapest, Hungary: Association
  for Computational Linguistics.

\bibitem[{Kumar et~al.(2021)Kumar, Fu, Pathak, and Malik}]{kumar2021rma}
Kumar, A.; Fu, Z.; Pathak, D.; and Malik, J. 2021.
\newblock {RMA}: Rapid Motor Adaptation for Legged Robots.
\newblock In \emph{Robotics: Science and Systems}.

\bibitem[{Lanier et~al.(2026)Lanier, Kim, Karamzade, Liu, Sinha, He, Corsi, and
  Fox}]{lanier2026redraw}
Lanier, J.~B.; Kim, K.; Karamzade, A.; Liu, Y.; Sinha, A.; He, K.; Corsi, D.;
  and Fox, R. 2026.
\newblock Adapting World Models with Latent-State Dynamics Residuals.
\newblock In \emph{Proceedings of the 8th Annual Learning for Dynamics and
  Control Conference}, volume 331 of \emph{Proceedings of Machine Learning
  Research}, 117--144.

\bibitem[{Miconi, Clune, and Stanley(2018)}]{miconi2018differentiable}
Miconi, T.; Clune, J.; and Stanley, K.~O. 2018.
\newblock Differentiable Plasticity: Training Plastic Neural Networks with
  Backpropagation.
\newblock In \emph{Proceedings of the 35th International Conference on Machine
  Learning}, volume~80 of \emph{Proceedings of Machine Learning Research},
  3559--3568.

\bibitem[{Nagabandi et~al.(2019)Nagabandi, Clavera, Liu, Fearing, Abbeel,
  Levine, and Finn}]{nagabandi2019learning}
Nagabandi, A.; Clavera, I.; Liu, S.; Fearing, R.~S.; Abbeel, P.; Levine, S.;
  and Finn, C. 2019.
\newblock Learning to Adapt in Dynamic, Real-World Environments Through
  Meta-Reinforcement Learning.
\newblock In \emph{International Conference on Learning Representations}.

\bibitem[{O'Connell et~al.(2022)O'Connell, Shi, Shi, Azizzadenesheli,
  Anandkumar, Yue, and Chung}]{oconnell2022neuralfly}
O'Connell, M.; Shi, G.; Shi, X.; Azizzadenesheli, K.; Anandkumar, A.; Yue, Y.;
  and Chung, S.-J. 2022.
\newblock {Neural-Fly} Enables Rapid Learning for Agile Flight in Strong Winds.
\newblock \emph{Science Robotics}, 7(66): eabm6597.

\bibitem[{Pannocchia and Rawlings(2003)}]{pannocchia2003disturbance}
Pannocchia, G.; and Rawlings, J.~B. 2003.
\newblock Disturbance Models for Offset-Free Model-Predictive Control.
\newblock \emph{{AIChE} Journal}, 49(2): 426--437.

\bibitem[{Richards et~al.(2023)Richards, Azizan, Slotine, and
  Pavone}]{richards2023control}
Richards, S.~M.; Azizan, N.; Slotine, J.-J.; and Pavone, M. 2023.
\newblock Control-oriented meta-learning.
\newblock \emph{The International Journal of Robotics Research}, 42(10):
  777--797.

\bibitem[{Schlag, Irie, and Schmidhuber(2021)}]{schlag2021linear}
Schlag, I.; Irie, K.; and Schmidhuber, J. 2021.
\newblock Linear Transformers Are Secretly Fast Weight Programmers.
\newblock In \emph{Proceedings of the 38th International Conference on Machine
  Learning}, volume 139 of \emph{Proceedings of Machine Learning Research},
  9355--9366.

\bibitem[{Schmidhuber(1992)}]{schmidhuber1992fast}
Schmidhuber, J. 1992.
\newblock Learning to Control Fast-Weight Memories: An Alternative to Dynamic
  Recurrent Networks.
\newblock \emph{Neural Computation}, 4(1): 131--139.

\bibitem[{Smith, Ghazizadeh, and Shadmehr(2006)}]{smith2006interacting}
Smith, M.~A.; Ghazizadeh, A.; and Shadmehr, R. 2006.
\newblock Interacting Adaptive Processes with Different Timescales Underlie
  Short-Term Motor Learning.
\newblock \emph{PLoS Biology}, 4(6): e179.

\bibitem[{Soni et~al.(2026{\natexlab{a}})Soni, Sehgal, Dave, and
  Rueckert}]{soni2026alra}
Soni, K.; Sehgal, A.; Dave, V.; and Rueckert, E. 2026{\natexlab{a}}.
\newblock {ALRA}: Action-Latent Residual Adaptation for World Models.
\newblock In \emph{NeurIPS Workshop on PTA: From Pretrained Representations to
  Acting Agents}.
\newblock \url{https://openreview.net/forum?id=r5DF75OXlT}.

\bibitem[{Soni et~al.(2026{\natexlab{b}})Soni, Sehgal, Dave, and
  Rueckert}]{soni2026sandwich}
Soni, K.; Sehgal, A.; Dave, V.; and Rueckert, E. 2026{\natexlab{b}}.
\newblock {Sandwich-Residuals}: Parameter-Efficient Test-time Adaptation of
  World Models.
\newblock \emph{arXiv preprint arXiv:2609.21740}.

\bibitem[{Tassa et~al.(2018)Tassa, Doron, Muldal, Erez, Li, de~Las~Casas,
  Budden, Abdolmaleki, Merel, Lefrancq, Lillicrap, and
  Riedmiller}]{tassa2018dmcontrol}
Tassa, Y.; Doron, Y.; Muldal, A.; Erez, T.; Li, Y.; de~Las~Casas, D.; Budden,
  D.; Abdolmaleki, A.; Merel, J.; Lefrancq, A.; Lillicrap, T.; and Riedmiller,
  M. 2018.
\newblock {DeepMind} Control Suite.
\newblock \emph{arXiv preprint arXiv:1801.00690}.

\bibitem[{Wang et~al.(2026)Wang, Bounou, LeCun, and Ren}]{wang2026adajepa}
Wang, Y.; Bounou, O.; LeCun, Y.; and Ren, M. 2026.
\newblock {AdaJEPA}: An Adaptive Latent World Model.
\newblock In \emph{NeurIPS Workshop on Continual World Models}.
\newblock {arXiv}:2606.32026.

\bibitem[{Williams, Aldrich, and Theodorou(2017)}]{williams2017mppi}
Williams, G.; Aldrich, A.; and Theodorou, E.~A. 2017.
\newblock Model Predictive Path Integral Control: From Theory to Parallel
  Computation.
\newblock \emph{Journal of Guidance, Control, and Dynamics}, 40(2): 344--357.

\bibitem[{Wolpert, Miall, and Kawato(1998)}]{wolpert1998internal}
Wolpert, D.~M.; Miall, R.~C.; and Kawato, M. 1998.
\newblock Internal Models in the Cerebellum.
\newblock \emph{Trends in Cognitive Sciences}, 2(9): 338--347.

\bibitem[{Zhang et~al.(2026)Zhang, Ye, Agarwal, {Nourkhiz Mahjoub}, {Moradi
  Pari}, Khashabi, Shu, and Tadiparthi}]{zhang2026jepattt}
Zhang, Z.; Ye, S.; Agarwal, N.; {Nourkhiz Mahjoub}, H.; {Moradi Pari}, E.;
  Khashabi, D.; Shu, T.; and Tadiparthi, V. 2026.
\newblock {JEPA-TTT}: Persistent Test-Time Training of Latent World Models for
  Planning under Dynamics Shifts.
\newblock In \emph{NeurIPS Workshop on World Models in Physical AI}.
\newblock {arXiv}:2610.00722.

\bibitem[{Zhou et~al.(2025)Zhou, Pan, LeCun, and Pinto}]{zhou2025dinowm}
Zhou, G.; Pan, H.; LeCun, Y.; and Pinto, L. 2025.
\newblock {DINO-WM}: World Models on Pre-trained Visual Features enable
  Zero-shot Planning.
\newblock In \emph{Proceedings of the 42nd International Conference on Machine
  Learning}, volume 267 of \emph{Proceedings of Machine Learning Research},
  79115--79135. PMLR.

\end{thebibliography}

\clearpage
\appendix

\section{Implementation Details}
\label{app:impl}

All primary controllers use the state-input TD-MPC2 checkpoint numbered 1 for their task. Action dimensions are 6 (Walker Walk), 4 (Hopper Hop), 6 (Cheetah Run), 5 (Fish Swim), and 2 (Reacher Hard). Every model has a 512-coordinate latent with SimNorm groups of eight. The key projection has 64 columns and stays fixed throughout.

Each episode resets the environment, the planner random state, the planner warm start, and the fast state. The frozen condition keeps the original transition read and performs no writes. Imagined rollouts hold the fast state fixed. No reward signal enters the update. The bias uses a Euclidean norm cap and the matrix a Frobenius norm cap, both of radius 10.

\paragraph{Local objective.}
For one SimNorm group, write $v=\log\hat z+\Delta$ and $y=\mathrm{softmax}(v)$. For a target $t$ that sums to one, differentiating $-\sum_i t_i\log y_i$ with respect to $v$ gives $y-t=-e$, and multiplying by $x^\top$ yields Eq.~\eqref{eq:local}.

\paragraph{Uncertainty.}
For seed-level paired differences $d_1,\dots,d_n$ we report $\bar d\pm t_{0.975,n-1}\,s_d/\sqrt n$. The confirmation cohort contains 800 episodes: for each of 20 seeds and five tasks, frozen and keyed at gain 1.0 and frozen, keyed, and bias at gains 0.7 and 0.6. A further 60 bias episodes at gain 1.0 on Walker, Hopper, and Cheetah complete Table~\ref{tab:main}. They are paired with the stored frozen episodes after six reruns (seeds 20 and 21 on each task) reproduced the stored frozen returns exactly. All episodes complete 500 decisions.

\paragraph{Timing.}
MPC planning takes 29.5--29.9~ms per decision with the bias read and 32.4--32.8~ms with the keyed read. The online gain estimator adds 50.5--50.8~ms.

\section{Constant-Key Reduction}
\label{app:constkey}

Let the key be a constant $x_t=c_x$ and define $b_t=P_t c_x$. Right-multiplying Eq.~\eqref{eq:keyed} by $c_x$ with an inactive cap gives
\begin{equation}
b_{t+1}=(1-\lambda)\,b_t+\eta\,\frac{\lVert c_x\rVert_2^2}{\lVert c_x\rVert_2^2+\epsilon}\,e_t ,
\end{equation}
which is Eq.~\eqref{eq:bias} for $\lVert c_x\rVert_2^2\gg\epsilon$. In the experiment, $c_x$ is estimated from an independent nominal Walker rollout (seed 10000). On Walker at gain 0.7 (seeds 0--19), frozen, keyed, constant-key, and bias controllers return 959.65, 969.49, 970.86, and 971.58.

\paragraph{Key geometry.}
Mean pairwise cosine similarity of the normalized keys over 200 decisions is 0.686 on Walker, 0.601 on Cheetah, and 0.700 on Fish. The fractions of pairs above 0.9 are 0.031, 0.032, and 0.056.

\section{Additional Control Results}

\subsection{Reacher Hard and Nominal Dynamics}
\label{app:reacher}

Reacher keeps high return under actuation changes: frozen means are 980.70, 973.80, and 969.60 at gains 1.0, 0.7, and 0.6. Keyed return differences are $0.00$, $-0.05$, and $0.00$, and bias differences are $0.00$ at both shifted gains. Same-transition prediction still improves, by $6.1\times10^{-5}$ (keyed) and $5.3\times10^{-5}$ (bias) in MSE at gain 0.7. Under alternating gains, mixed gains, and 30\% higher hinge damping, all paired differences lie within 0.2 points of zero.

At nominal gain on seeds 20--39, final mean keyed Frobenius norms are 0.48, 1.02, and 0.93 for Walker, Hopper, and Cheetah, and final mean bias norms are 0.128, 0.280, and 0.205. No episode of either form activates the cap. Mean same-transition MSE improvements for the bias are $3.06\times10^{-7}$, $3.40\times10^{-6}$, and $1.24\times10^{-6}$. The two norms describe different parameterizations and are not compared directly.

\subsection{Mass Change on All Bodies}
\label{app:mass}

\begin{table}[h]
\centering
\small
\setlength{\tabcolsep}{3pt}
\resizebox{\columnwidth}{!}{\begin{tabular}{lrrr}
\toprule
Task & Frozen & Keyed $\Delta$ [95\% CI] & Bias $\Delta$ [95\% CI] \\
\midrule
Walker & 984.64 & $0.04$ [$-0.48$, $0.55$] & $-0.11$ [$-0.75$, $0.54$] \\
Hopper & 242.75 & $44.98$ [$34.49$, $55.48$] & $52.80$ [$43.00$, $62.59$] \\
Cheetah & 823.56 & $-41.52$ [$-128.03$, $45.00$] & $-62.55$ [$-206.25$, $81.15$] \\
Fish & 805.03 & $-16.10$ [$-52.58$, $20.37$] & $6.18$ [$-10.65$, $23.01$] \\
\bottomrule
\end{tabular}}
\caption{Torso mass $\times 1.3$ on seeds 20--39 (20 pairs per row).}
\label{tab:mass}
\end{table}

Table~\ref{tab:mass} gives the mass change on every body. Hopper is the body whose return the mass change lowers substantially, and it is where plasticity recovers it.

\subsection{Development Cohort and Walker Gain Sweep}

On seeds 0--19 at gain 0.7, keyed gains are 9.84, 13.77, and 38.00 points for Walker, Hopper, and Cheetah, the same direction as the confirmation cohort. Table~\ref{tab:sweep} extends the Walker sweep. Under mixed joint gains, keyed plasticity raises Cheetah return by 22.68 points.

\begin{table}[h]
\centering
\small
\begin{tabular}{rrrrr}
\toprule
Gain & Frozen & Keyed & Mean diff. & Paired SD \\
\midrule
0.50 & 264.95 & 318.53 & 53.58 & 116.38 \\
0.60 & 708.04 & 860.22 & 152.18 & 175.45 \\
0.70 & 959.65 & 969.49 & 9.84 & 11.34 \\
0.85 & 983.12 & 981.98 & $-1.13$ & 5.17 \\
1.00 & 983.88 & 983.34 & $-0.55$ & 3.06 \\
\bottomrule
\end{tabular}
\caption{Walker gain sweep on development seeds 0--19.}
\label{tab:sweep}
\end{table}

\subsection{Additional Pretrained Checkpoints}
\label{app:ckpt}

\begin{table}[h]
\centering
\small
\setlength{\tabcolsep}{4pt}
\begin{tabular}{llrrr}
\toprule
Task / ckpt. & Rule & 0--9 & 20--29 & 30--39 \\
\midrule
Hopper / 2 & Keyed & $25.38$ & $28.99$ & $17.78$ \\
 & Bias & $34.24$ & $26.28$ & $20.32$ \\
Hopper / 3 & Keyed & $0.93$ & $0.52$ & $1.73$ \\
 & Bias & $2.29$ & $0.03$ & $0.32$ \\
Cheetah / 2 & Keyed & $64.13$ & $27.76$ & $-13.42$ \\
 & Bias & $118.94$ & $22.33$ & $45.79$ \\
Cheetah / 3 & Keyed & $15.89$ & $32.26$ & $40.54$ \\
 & Bias & $131.90$ & $70.88$ & $90.53$ \\
\bottomrule
\end{tabular}
\caption{Mean paired return gain at gain 0.7 for additional pretrained checkpoints, in three ten-seed blocks.}
\label{tab:ckpt}
\end{table}

Table~\ref{tab:ckpt} reports checkpoints 2 and 3 for Hopper and Cheetah. Of the 24 block means, 23 are positive. Hopper checkpoint~2 improves in every block with a first-block interval of $[19.36, 31.41]$ for the keyed state. Hopper checkpoint~3 sits near the reward floor and changes little. Cheetah's ten-seed intervals are wide.

\subsection{Explicit Gain Models}

The estimator evaluates gains in $\{0.4,0.5,\dots,1.0\}$ against up to 25 recent transitions and applies the selected gain to the dynamics input in the next planning call. On Cheetah its mean final estimates are 0.705 and 0.605 at gains 0.7 and 0.6. On Hopper at gain 0.7 the mean estimate is 0.700 and six of 20 episodes end on a different gain. An inverse-command reference rescales the issued action by the true gain. It returns 975.82 on Walker at gain 0.7 and 443.60 and 292.63 on Cheetah at gains 0.7 and 0.6. Its commands clip on at least 94\% of decisions.

\section{Further Controls and Sensitivity}
\label{app:controls}

\paragraph{Integration controls.}
Carrying forward the previous log-latent residual without a leaky error-driven update changes return by $-208.61$ on Walker at gain 0.7 and $-134.74$ on Cheetah at gain 0.6 (seeds 0--19). A diagonal Kalman bias gains $+14.91$ and $+26.86$ on the same settings.

\paragraph{Sparse expanded key.}
The expanded key replaces the 64-dimensional projection with a 512-dimensional ReLU projection that keeps its 51 largest coordinates, with the write rule unchanged. On Walker at gain 0.7 (seeds 0--19) paired gains are $8.48$ for the expanded key, $9.84$ for the original key, and $11.93$ for the bias. On Fish the expanded key has a mean gain of $17.31$ $[-15.63, 50.24]$.

\paragraph{Update parameters.}
Table~\ref{tab:sens} varies one parameter at a time. Doubling $\eta$ to 0.10 raises mean return in every block. The effect of the leak differs between seed blocks. We kept the default throughout.

\begin{table}[h]
\centering
\small
\begin{tabular}{llrrr}
\toprule
$\eta$ & $\lambda$ & W 0--9 & W 10--19 & C 0--19 \\
\midrule
0.025 & 0.001 & $-11.83$ & $2.49$ & $19.46$ \\
0.050 & 0.001 & $9.82$ & $9.86$ & $19.68$ \\
0.050 & 0 & $-4.73$ & $13.48$ & $29.32$ \\
0.050 & 0.010 & $-0.95$ & $4.99$ & $21.23$ \\
0.100 & 0.001 & $14.21$ & $17.37$ & $25.91$ \\
\bottomrule
\end{tabular}
\caption{Mean paired return gain of keyed plasticity for Walker at gain 0.7 (W) and Cheetah at gain 0.6 (C) by seed block.}
\label{tab:sens}
\end{table}

\paragraph{Restoring nominal dynamics.}
We ran gain 1.0 for decisions 0--99, gain 0.7 for decisions 100--299, and gain 1.0 again for decisions 300--499 on ten Walker seeds. Returns over the final 200 decisions are 399.85 of a possible 400 whether the adapted state is retained, decayed, or reset. On the first restored transition the retained state has latent MSE $4.22\times10^{-5}$ against $3.48\times10^{-5}$ for a zero state.

\section{Fish Swim}
\label{app:fish}

\begin{table}[h]
\centering
\small
\setlength{\tabcolsep}{3.5pt}
\begin{tabular}{rrrrrr}
\toprule
Gain & $\alpha$ & Cap & Return & $\Delta R$ [95\% CI] & Cap ep. \\
\midrule
0.7 & 1 & 10 & 633.26 & $-23.94$ [$-61.89$, $14.01$] & 18/20 \\
0.7 & 0.5 & 10 & 652.43 & $-4.77$ [$-40.11$, $30.56$] & 19/20 \\
0.7 & 0.25 & 10 & 654.96 & $-2.25$ [$-23.70$, $19.20$] & 20/20 \\
0.7 & 1 & 5 & 649.40 & $-7.80$ [$-28.19$, $12.59$] & 20/20 \\
0.7 & 1 & 3 & 665.15 & $7.94$ [$-14.24$, $30.12$] & 20/20 \\
0.6 & 1 & 10 & 504.27 & $-27.13$ [$-142.68$, $88.43$] & 18/20 \\
0.6 & 0.5 & 10 & 550.42 & $19.03$ [$-73.75$, $111.81$] & 19/20 \\
0.6 & 0.25 & 10 & 554.92 & $23.52$ [$-29.21$, $76.26$] & 20/20 \\
0.6 & 1 & 5 & 540.83 & $9.44$ [$-93.59$, $112.46$] & 20/20 \\
0.6 & 1 & 3 & 512.27 & $-19.13$ [$-117.13$, $78.88$] & 20/20 \\
\bottomrule
\end{tabular}
\caption{Fish Swim with the keyed read scaled by $\alpha$ and the norm cap reduced (development seeds 0--19).}
\label{tab:fish}
\end{table}

Table~\ref{tab:fish} scales the keyed read as $\SimNorm(\log\hat z+\alpha P x)$ and varies the norm cap. Weaker reads move mean return up at both gains. Under structured changes on the same cohort, keyed plasticity gains $16.12$ $[5.88, 26.35]$ under alternating gains with no cap activation, changes return by $-85.98$ $[-171.45, -0.51]$ under mixed gains with the cap active in 19 of 20 episodes, and gains $8.43$ $[-2.14, 19.01]$ under a 30\% mass increase. The bias changes return by $2.44$, $-120.55$, and $-0.87$ on the same three changes. The pattern matches the main text: small corrections help or are neutral, and large corrections hurt.

\end{document}